\documentclass[runningheads]{llncs}

\usepackage[cameraready,year=2025,ID=3]{iciap}
\usepackage{iciap}

\usepackage{iciapabbrv}
\usepackage{adjustbox}
\usepackage{graphicx}
\usepackage{booktabs}
\usepackage{subcaption}

\usepackage[accsupp]{axessibility}  
 \usepackage{array,multirow,graphicx}
 \usepackage{float}

\newcommand{\methodname}{SISMA}

\usepackage[pagebackref,breaklinks,colorlinks,citecolor=iciapblue]{hyperref}

\usepackage{orcidlink}

\begin{document}

\title{\methodname{}:  \underline{S}emantic Face \underline{I}mage \underline{S}ynthesis with \underline{Ma}mba} 

\titlerunning{\methodname{}}

\author{Filippo Botti\inst{1}\orcidlink{0009-0001-7567-753X} \and
Alex Ergasti\inst{1}\orcidlink{0009-0005-8110-9714} \and
Tomaso Fontanini\inst{1}\orcidlink{0000-0001-6595-4874} \and
Claudio Ferrari\inst{2}\orcidlink{0000-0001-9465-6753} \and
Massimo Bertozzi\inst{1}\orcidlink{0000-0003-1463-5384} \and
Andrea Prati\inst{1}\orcidlink{0000-0002-1211-529X} 
}

\authorrunning{F.~Botti et al.}

\institute{University of Parma, Department of Engineering and Architecture, 43124 Parma, Italy    
\and
University of Siena, Department of Information Engineering and Mathematical Sciences, 53100 Siena, Italy
\\ \email{filippo.botti@unipr.it} 
}

\maketitle

\begin{abstract}
  Diffusion Models have become very popular for Semantic Image Synthesis (SIS) of human faces. Nevertheless, their training and inference is computationally expensive and their computational requirements are high due to the quadratic complexity of attention layers. In this paper, we propose a novel architecture called \methodname{}, based on the recently proposed Mamba. \methodname{} generates high quality samples by controlling their shape using a semantic mask at a reduced computational demand. We validated our approach through comprehensive experiments with CelebAMask-HQ, revealing that our architecture not only achieves a better FID score yet also operates at three times the speed of state-of-the-art architectures. This indicates that the proposed design is a viable, lightweight substitute to transformer-based models.
\end{abstract}

\section{Introduction}

Semantic Image Synthesis (SIS) refers to the task of generating samples conditioned on a semantic mask, that is an image in which each pixel identifies a specific semantic class. In the case of human faces, each class define a different part such as eyes, hair or mouth. The idea is that the mask can condition the shape of the generated samples, while the styles and textures of the semantic parts are either inferred by the generative model, or extracted by a target image. 

In the past years, SIS was dominated by methods based on Generative Adversarial Networks (GANs), with several solutions that were proposed usually employing custom normalization layers to inject the mask information. Among them, one of the most popular is SEAN \cite{sean} which can both control the shape and the style of the generated samples with high precision. Nevertheless, SEAN can not freely generate samples without a supporting style image, and therefore lacks diversity in the generation. Other approaches like \cite{shi2022semanticstylegan} can instead generate highly diverse samples, but lack fine-grained control over the style of specific semantic classes. Taking the previous methods as an example, we can therefore divide SIS architectures in two categories: \textit{style-driven} approaches, in which a reference image is always necessary to guide the style of the generated samples, and \textit{diversity-driven} approaches, in which the style is generated from scratch without any specific reference. Despite this coarse categorization, some recent works showed that performing both is also possible~\cite{clade}.  

Thanks to the great success of Diffusion Models (DM) in various generation tasks, and especially in text-to-image generation \cite{ldm}, they have recently been utilized also in SIS. Several DM-based methods have been proposed such as \cite{SDM,cnet,cdiffusion}, which reached impressive results in diversity-driven generation. Nevertheless, the major drawback of DMs is the huge amount of resources required for training. Performing inference is also a costly process, which limits their practical use. For this reason, recently, State Space Models (SSM) \cite{gu2021efficiently,gu2021combining} have attracted a lot of attention due to their great efficiency. In particular, Mamba \cite{gu2023mamba} became popular due to its capability of competing with transformer-based architectures while requiring much less computational resources. This is due to the fact that the memory required by Mamba grows linearly with respect to the sequence length rather than quadratically as in transformers. Due to the aforementioned advantages, many Mamba-based diffusion models have been developed \cite{hu2024zigma, ergasti2025ushapemambastatespace, gao2024mattenvideogenerationmambaattention}.

In this paper, we propose \methodname{}, a novel diffusion model architecture based on Mamba specifically tailored for SIS. In particular, our focus was on the generation of human faces, since the possibility of controlling the shape when synthesizing new identities is of paramount importance for several applications such as generating different variations of the same face. Our architecture substitutes the transformer blocks of DiT \cite{peebles2023scalable} with novel \methodname{} blocks that can be conditioned by semantic masks during the generation without the need of custom normalization or attention layers. In this work we provide the following contributions:
\begin{itemize}
    \item We designed and propose an architecture based on Mamba that can perform diversity-driven SIS with greater efficiency but comparable quality. 
    \item A novel block, named \methodname{}, that combines a Self-Mamba layer to model intra-feature interactions and a Cross-Mamba layer that enables the mask conditioning to control the shape of the generated samples.
\end{itemize}

\section{Related Work}

Semantic image synthesis (SIS) architectures can be broadly divided into two main categories: those based on Generative Adversarial Networks (GANs) and those leveraging diffusion models. Within the GAN-based paradigm, methods are further distinguished by whether or not they rely on reference style images.

Among the GAN-based approaches, SEAN~\cite{sean} stands out for its use of SPatial Adaptive DE-normalization (SPADE) layers~\cite{spade}, which inject semantic segmentation masks and style embeddings into the generator. However, SEAN suffers from limited output diversity, producing identical results for identical inputs. Another notable model is CLADE (CLass-Adaptive DE-normalization) \cite{clade}, which supports both style-driven and diversity-driven synthesis. This flexibility makes it suitable for a broader range of applications. Additionally, INADE \cite{inade} takes a different approach by conditioning on instance-level style information instead of class-level, offering finer control over individual elements in the generated image. Meanwhile, FrankenMask~\cite{fontanini2023frankenmask} and its variant~\cite{fontanini2023automatic} pushes the boundaries of automated semantic mask manipulation, further extending the capabilities of SIS. SemanticStyleGAN~\cite{shi2022semanticstylegan} represents a novel evolution of the StyleGAN~\cite{karras2019stylebased} family. Unlike earlier models, it separates style embeddings from semantic mask information and avoids direct conditioning on the mask. It employs GAN inversion~\cite{zhu2020domain} to generate facial images from semantic layouts. Additionally, Tarollo \textit{et al.}~\cite{tarollo2024adversarial} showcased a SIS model based on \cite{fontanini2025semantic} that can be used in adversarial attacks against facial recognition systems, underlining their broader relevance.

In the diffusion model domain, the Semantic Diffusion Model (SDM)~\cite{SDM} has gained attention for its simplicity and effectiveness. It can generate diverse images conditioned solely on semantic masks, though it lacks the control provided by reference images. Other diffusion-based techniques include ControlNet~\cite{cnet}, Composer~\cite{huang2023composer}, and T2I~\cite{t2i}, all of which enhance synthesis control by injecting semantic mask features into UNet architectures. Collaborative Diffusion~\cite{cdiffusion} introduces a multi-model cooperative setup, producing high-quality facial images. Finally, SCA-DM~\cite{ergasti2024controllablefacesynthesissemantic} is one of the latest models in the field, supporting both style-aware and style-agnostic semantic synthesis.

Differently from the existing literature, in this work we leverage the efficiency of state space models to propose an alternative SIS diffusion model with linear complexity that do not necessitate of any attention mechanism or custom normalization techniques to condition the generation.

\section{Methodology}
Our proposed architecture is based on Mamba, and is trained following a flow-matching strategy. To make the paper self-contained, we first briefly describe these two approaches in the following sections before delving into \methodname{}.

\subsection{Flow Matching}
Flow Matching (FM) \cite{liu2022flow,lipman2023flow} has been proposed as a simplified version of the diffusion forward process. It defines the latter as a linear interpolation between a sample from the data distribution $x \sim p_{data}$ and one from a Gaussian distribution $\varepsilon \sim \mathcal{N}(0,1)$. The interpolation can be defined depending on a variable $t$:
\begin{equation}
    z=tx+(1-t)\varepsilon, \quad t\in[0,1]
    \label{eq:fp}
\end{equation}

\noindent The backward process can then be solved using an ordinary differential equation:
\begin{equation}
    \frac{dz}{dt}=v=x-\varepsilon
\end{equation}

\noindent The model is thus trained to estimate the velocity $v$ with $v_\theta(z,t)$ (\textit{i.e.}, the derivative of Eq.~\ref{eq:fp}) taking as input both $z$ and $t$.

\begin{equation}
    \mathcal{L}=||(x-\varepsilon)-v_\theta(z,t)||^2
\end{equation}
%

\subsection{Mamba}
Mamba is a sequence-to-sequence architecture based on SSMs \cite{gu2023mamba,dao2024transformers}. It is designed to transform an input sequence $x(t) \in \mathbb{R}$ into an output sequence $y(t) \in \mathbb{R}$ through the use of an internal state $h(t) \in \mathbb{R}^N$, with $N$ representing the dimensionality of the hidden state. The inner equations that describe Mamba, and SSMs, follow the ordinary differentiable equation (ODE) linear system: 
\begin{equation}
     \begin{split}
     h'(t) &= A h(t) + B x(t)  \\        
     y(t) &= C h(t) + D x(t) 
     \end{split}
     \label{eq:mamba}
\end{equation}
\noindent where the matrices $A \in \mathbb{R}^{N\times N},B\in \mathbb{R}^{N\times 1},C\in \mathbb{R}^{1\times N} $ and $ D\in \mathbb{R}$ are learnable. Given that the equations are defined over continuous time, discretization is necessary for incorporation into a deep learning framework. In accordance with the approach in \cite{gu2023mamba}, we utilize the zero-order holder (ZOH) method. Here, $\Delta$ represents the step size, while $\Bar{A}$ and $\Bar{B}$ denote the discretized matrices. The equations can thus be expressed in the form of a Recurrent Neural Network (RNN):
\begin{equation}
     \begin{split}
     h_{k} &= \Bar{A}h_{k-1} + \bar{B}x_k  \\        
     y_{k} &= Ch_k + Dx_k
     \end{split}
     \label{eq:mamba_disc}
\end{equation}
Lastly, a key insight is that in Mamba, unlike other SSMs, the matrices are both learnable and vary with the input. Consequently, for a given input $x$, the matrices $B, C$, and $\Delta$ are derived through a linear fully-connected layer as follows:
\begin{equation}
     \begin{split}
     B = \mathrm{Lin_B}(x), C = \mathrm{Lin_C}(x), \Delta = \mathrm{Lin_{\Delta}}(x)
     \end{split}
     \label{eq:mamba_input_dependency}
\end{equation}

\subsection{\methodname{}}

\begin{figure}[t]
    \centering
    \includegraphics[width=0.8\textwidth]{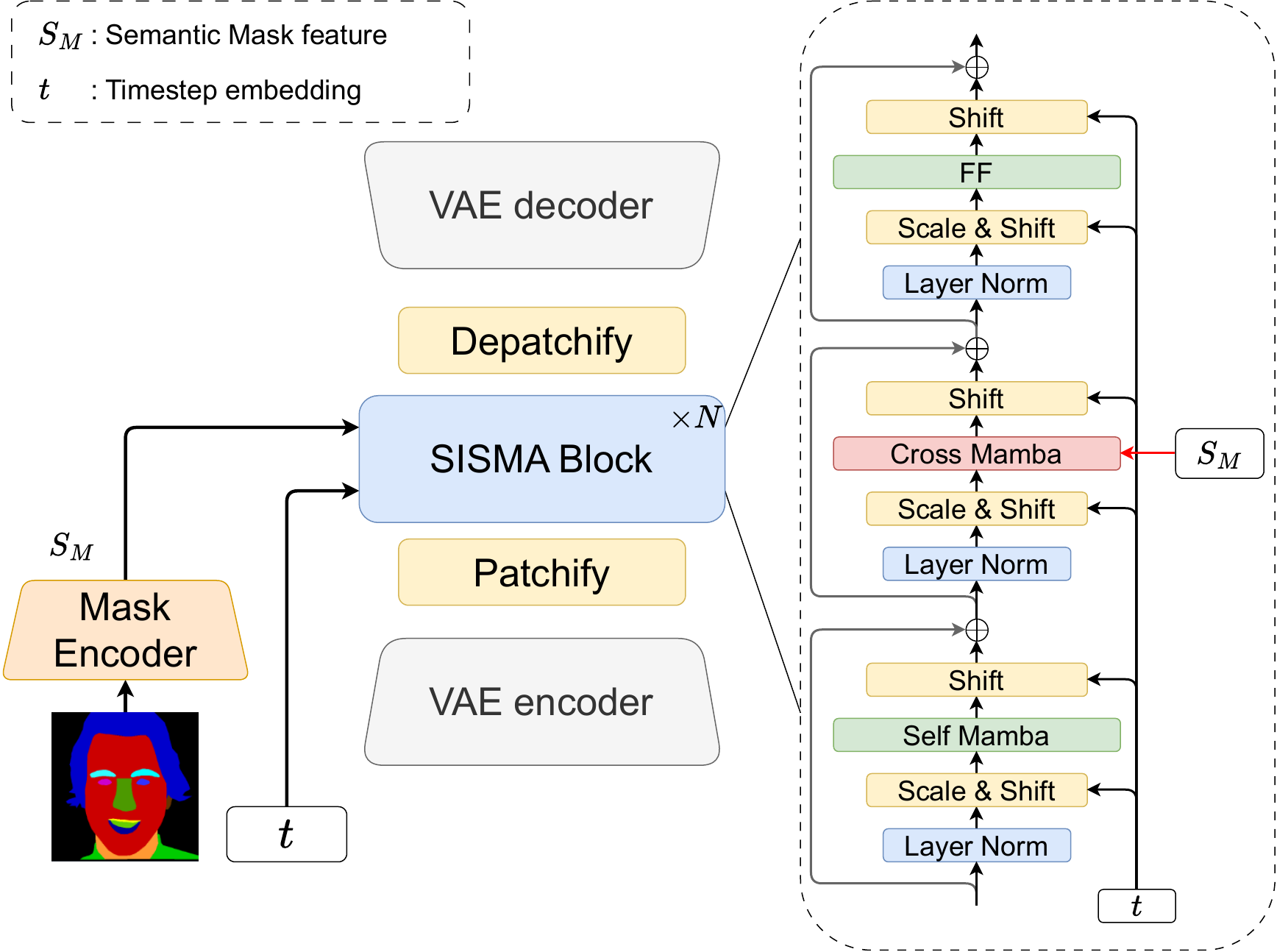}
    \caption{An overview of our architecture. A VAE encoder is used to reduces the input dimensionality, a Mask Encoder, trained with the model, is used to process the Semantic Mask, obtaining $S_M$. The core of the model are the $N$ \methodname{} blocks which are conditioned with $S_M$ and the diffusion timestep $t$.}
    \label{fig:arch}
\end{figure}

\begin{figure}
    \centering
    \begin{subfigure}{0.49\textwidth}
        \includegraphics[width=\textwidth]{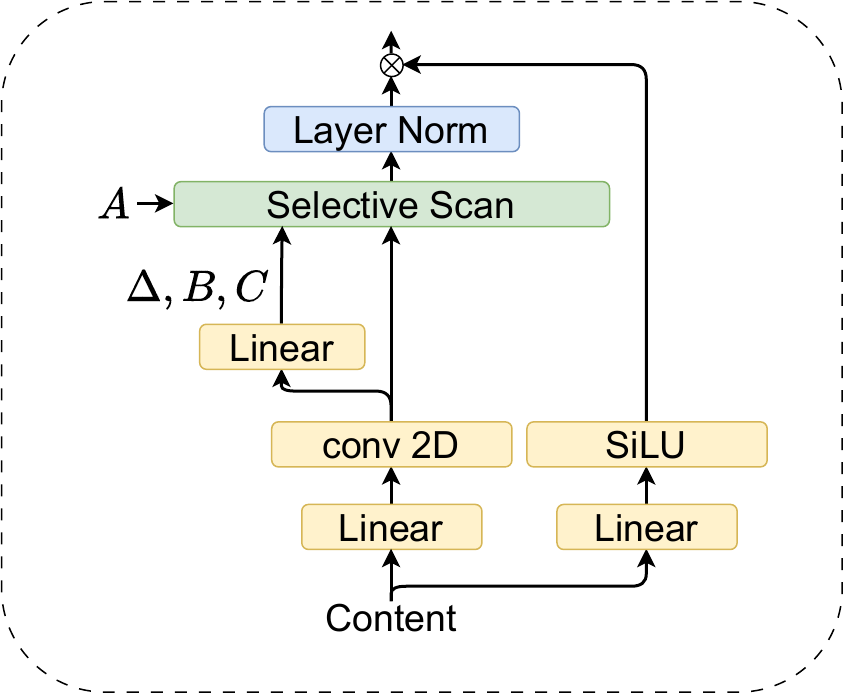}
        \caption{Self Mamba}
        \label{fig:self-mamba}
    \end{subfigure}
    \hfill
    \begin{subfigure}{0.49\textwidth}
        \includegraphics[width=\textwidth]{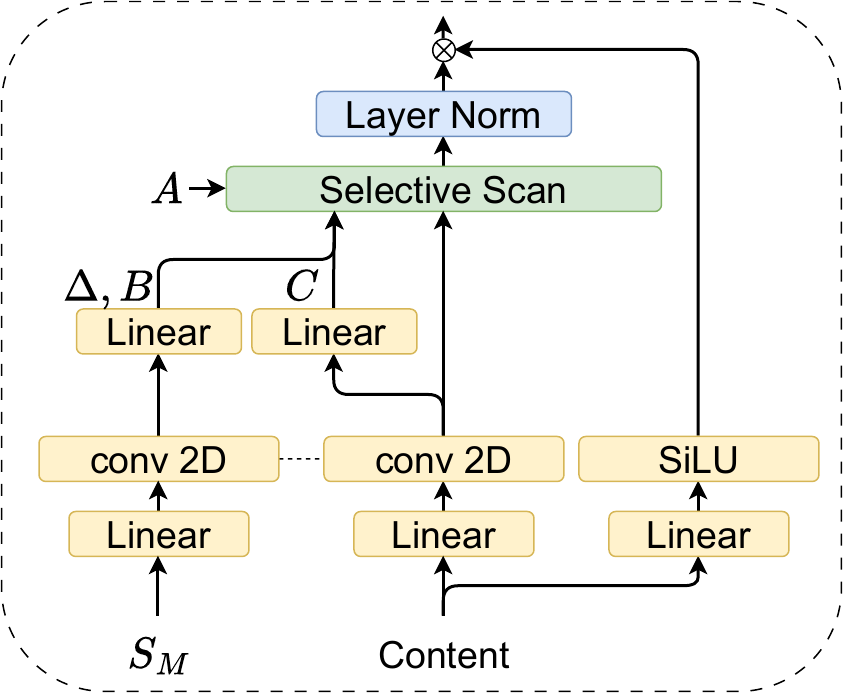}
        \caption{Cross Mamba}
        \label{fig:cross-mamba}
    \end{subfigure}
    \caption{a) An illustration of the Self Mamba, which is a basic mamba block. b) An illustration of Cross Mamba, in this block the matrices $\Delta$ and $B$ are obtained from the Semantic Mask $S_M$ in order to condition the diffusion process.}
    \label{fig:enter-label}
\end{figure}
Our architecture, called \methodname{}, illustrated in Fig.~\ref{fig:arch}, consists of several components. The encoder part of a pre-trained Variational AutoEncoder (VAE) \cite{ldm} is first used to reduce the input dimensionality. Given an image $I\in \mathbb{R}^{C\times H\times W}$, it maps $I$ to a new image $\hat{I}\in\mathbb{R}^{c\times h\times w}$, where $c=4$, $h=H/f$ and $w=W/f$, with $f$ being the scaling factor of the VAE. Then, a patchify layer is used to decompose $\hat{I}$ into a set of patches $x\in\mathbb{R}^{L\times D}$, where $D$ is the hidden size and $L=\frac{hw}{p^2}$, with $p$ being the patch size. Finally, $x$ is fed into the model which is composed of a series of \methodname{} blocks. In parallel, a Mask Encoder is used to process the Semantic Mask $m$ and produce a mask embedding $S_M\in\mathbb{R}^{L\times D}$, which is fed to each \methodname{} block in order to control the shape of the generated samples. The mask encoder is a simple encoder (a single conv2d layer with kernel $1 \times 1$) which map each class, passed as a one-hot mask, to a class embedding of size $\mathbb{R}^D$, obtaining an image $\in\mathbb{R}^{D\times h\times w}$ which is then pachified to obtain $S_M\in\mathbb{R}^{L\times D}$. Only the \methodname{} blocks and the Mask Encoder are trained, while the VAE encoder and decoder weights are frozen.

The core of our model comprises $N$ \methodname{} blocks, each conditioned on both $S_M$ and the diffusion timestep $t$. Each \methodname{} block contains three sequential components: (1) a Self-Mamba module that facilitates intra-feature interactions; (2) a Cross-Mamba module to explicitly condition the diffusion process on $S_M$; and (3) a Feed-Forward layer. Throughout the architecture, AdaLN \cite{peebles2023scalable} (Adaptive Layer Normalization) layers scale and shift features based on the timestep $t$ conditioning. After processing through the \methodname{} blocks, a depatchify operation is applied, which reverts the patchify layer, transforming the embedding from $\mathbb{R}^{L\times D}$ to $\mathbb{R}^{c\times h \times w}$. Lastly, a VAE decoder generates the final output. 

This design effectively integrates selective state-space models with conditional diffusion processes, allowing for semantic-guided generation through the specialized Cross Mamba conditioning mechanism \cite{10944183}. In detail, the Self-Mamba (Fig.~\ref{fig:self-mamba}) module leverages the selective scan mechanism to model sequential dependencies within the feature representation. On the other side, the Cross-Mamba (Fig.~\ref{fig:cross-mamba}) module extends this capability by incorporating conditioning from the semantic mask $S_M$, allowing for guided generation based on the provided semantic layout. Specifically, in Cross-Mamba, the selection matrices $\Delta$ and $B$ are computed from $S_M$, enabling fine-grained control over the generated content based on semantic regions, while the matrix $C$ is obtained from the content $x$, differently from equation \ref{eq:mamba_input_dependency}:
\begin{equation}
\begin{split}
 B = \mathrm{Lin_B}(S_M), \Delta = \mathrm{Lin_{\Delta}}(S_M), C = \mathrm{Lin_C}(x)
 \end{split}
 \label{eq:mamba_input_style_dependency}  
\end{equation}
By adopting this matrix generation, we are able to condition the generation phase with a semantic mask and obtain an image where the shape of the generated object follow the condition mask. 

\section{Results}

\subsection{Datasets}

We trained our model on CelebAMask-HQ \cite{celeba} dataset, which contains 30k images of human faces paired with the corresponding semantic mask. Each mask is composed by 19 different classes describing the shape of each face part, such as nose, hair, \textit{etc}...

\subsection{Training details}
In order to train our model, we use a batch size of 8, with a learning rate of 0.004 and no weight decay. Our architecture is composed of 24 \methodname{} blocks, each with a hidden dimension set to 1024. We do not perform any kind of data augmentation, except for a center cropping transformation and set $256\times 256$ as image size. In line with standard approaches in generative modeling research, we use an exponential moving average (EMA) with a decay factor of 0.9999 to track the weights throughout the training process. All the results and metrics are obtained with EMA model trained for 150K iteration. Finally, AdamW with default configuration \cite{kingma2014adam, loshchilov2017decoupled} is used as optimizer.

\subsection{Quantitative Results}
\begin{table*}[t]
\centering
\begin{minipage}{.49\textwidth}
\centering
\caption{Quantitative comparison with existing methods on semantic image synthesis. Best results are shown in \textbf{bold}. 0s in LPIPS means that the model cannot produce diverse results while - indicates that results for that dataset are not available.}
\adjustbox{width=\textwidth}{
\begin{tabular}{lcccc}
\toprule
{Method}
& FID $\downarrow$ & LPIPS $\uparrow$ \\
\midrule
    Pix2PixHD \cite{wang2018high} & 38.5 & 0 & \\
    SPADE \cite{spade}  & 29.2 & 0     \\
    DAGAN \cite{tang2020dual} & 29.1 & 0       \\
    SCGAN \cite{wang2021image} & 20.8 & 0    \\
    CLADE \cite{clade}  & 30.6 & 0   \\
    CC-FPSE \cite{liu2019learning} &   - & -   \\
    GroupDNet \cite{zhu2020semantically}  & 25.9 & 0.37   \\
    INADE \cite{inade}  & 21.5 & 0.415 \\
    SDM   \cite{SDM}    & 18.8 & 0.42   \\
\midrule
    \textbf{SISMA (Ours)}& \textbf{18.3} & \textbf{0.61} \\
\bottomrule
\end{tabular}
}
\label{tab:fid}
\end{minipage}
\hfill
\begin{minipage}{0.49\textwidth}
\centering
\caption{Efficiency of SDM in comparison with \methodname{}.}
\begin{tabular}{c|c|c}
\toprule
              & Params (M) & Inference Time (s) \\
\midrule
     SDM      & \textbf{653} & 22.79  \\
     SISMA    & 799 & \textbf{8.51} \\
\bottomrule
\end{tabular}
\label{tab:time}
\end{minipage}
\end{table*}

In Table \ref{tab:fid} comparisons with several state-of-the-art methods are presented. We evaluated the quality of the generated samples using the Frechet Inception Distance (FID), which measures the distance between the distribution of real images and the distribution of fake images, and the diversity in the generation using the Learned Perceptual Image Patch Similarity (LPIPS), which measures the similarity between two generated images. \methodname{} achieves results that are comparable with the state of the art, but without the need for custom normalization or attention layers. 
Additionally, regarding the diversity of the generated results, SISMA is able to outperform every other model, achieving an LPIPS of 0.61. This demonstrates our model's enhanced ability to generalize and produce different results from the same semantic mask.

Finally, thanks to its design, \methodname{} is much more efficient than other SIS diffusion models that employ custom normalization layers or attention layers like SDM, which makes use of both SPADE layers and self-attention layers. We verify this aspect by performing a runtime analysis reported in Table \ref{tab:time}. Despite having slightly higher number of parameters, \methodname{} generates samples 3 times faster than SDM on an NVIDIA RTX 4090. In particular, we measured the inference time to generate one image using 200 diffusion steps. Faster inference times imply our model can generate samples with a fraction of the cost with respect to vanilla diffusion models, which is crucial in such computational expensive architectures. 

Previous GAN-based approaches such as CLADE, INADE etc. are not considered in this comparison as their structure significantly differs from diffusion models. GANs are way faster at inference given that images are generated in a single, forward pass. However, diffusion models retain much higher quality and versatility; thus, we deem important to compare SISMA with previous diffusion-based approaches, with the goal of improving the computational aspect of such methods, which still represents a non-negligible bottleneck in their practical use.
\begin{figure}[H]
    \centering
    \includegraphics[width=\linewidth]{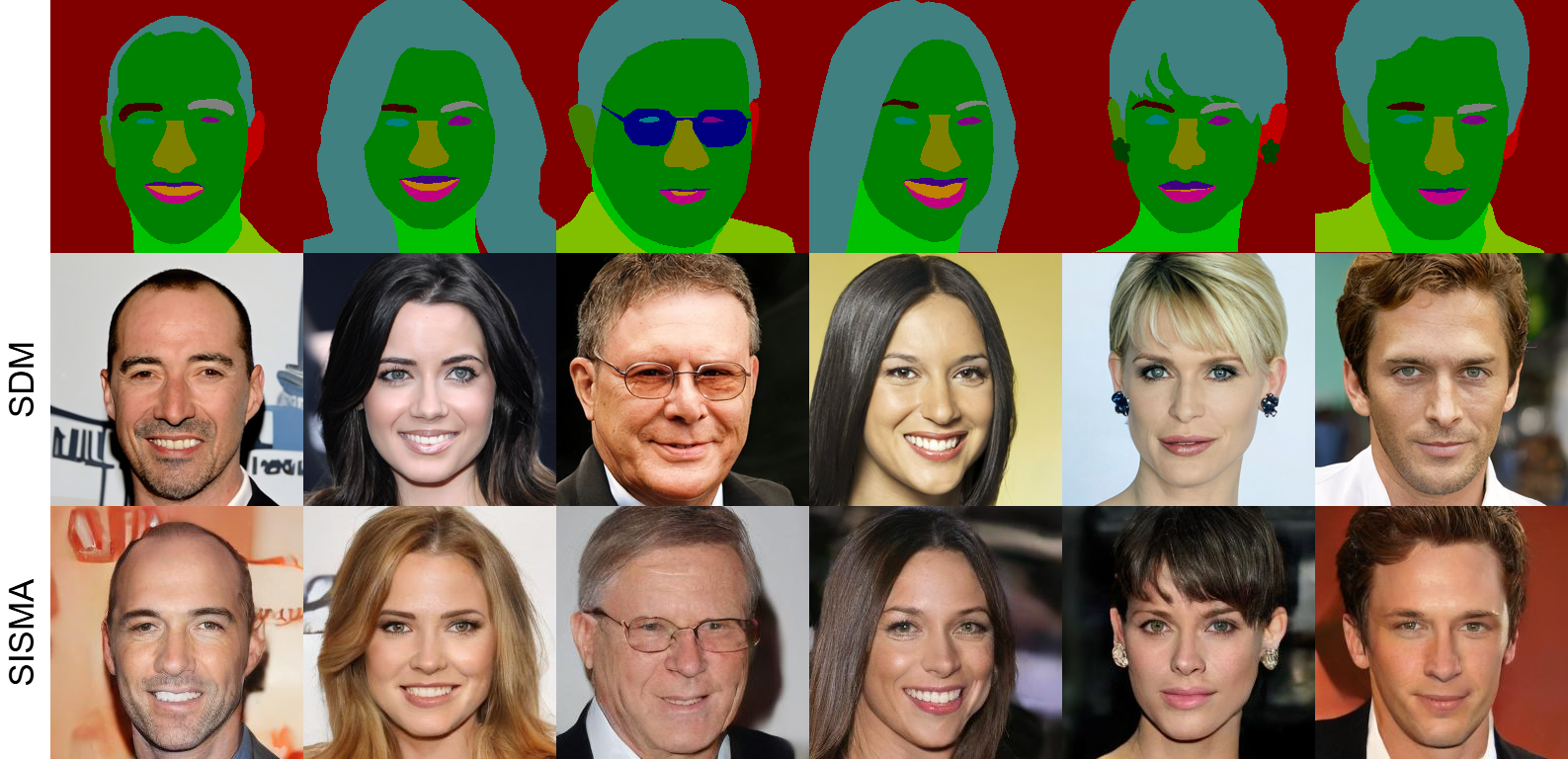}
    \caption{Comparison between SDM (second row) and SISMA (last row) on CelebA.}
    \label{fig:celeba}
\end{figure}
\begin{figure}[H]
    \centering
    \includegraphics[width=0.9\linewidth]{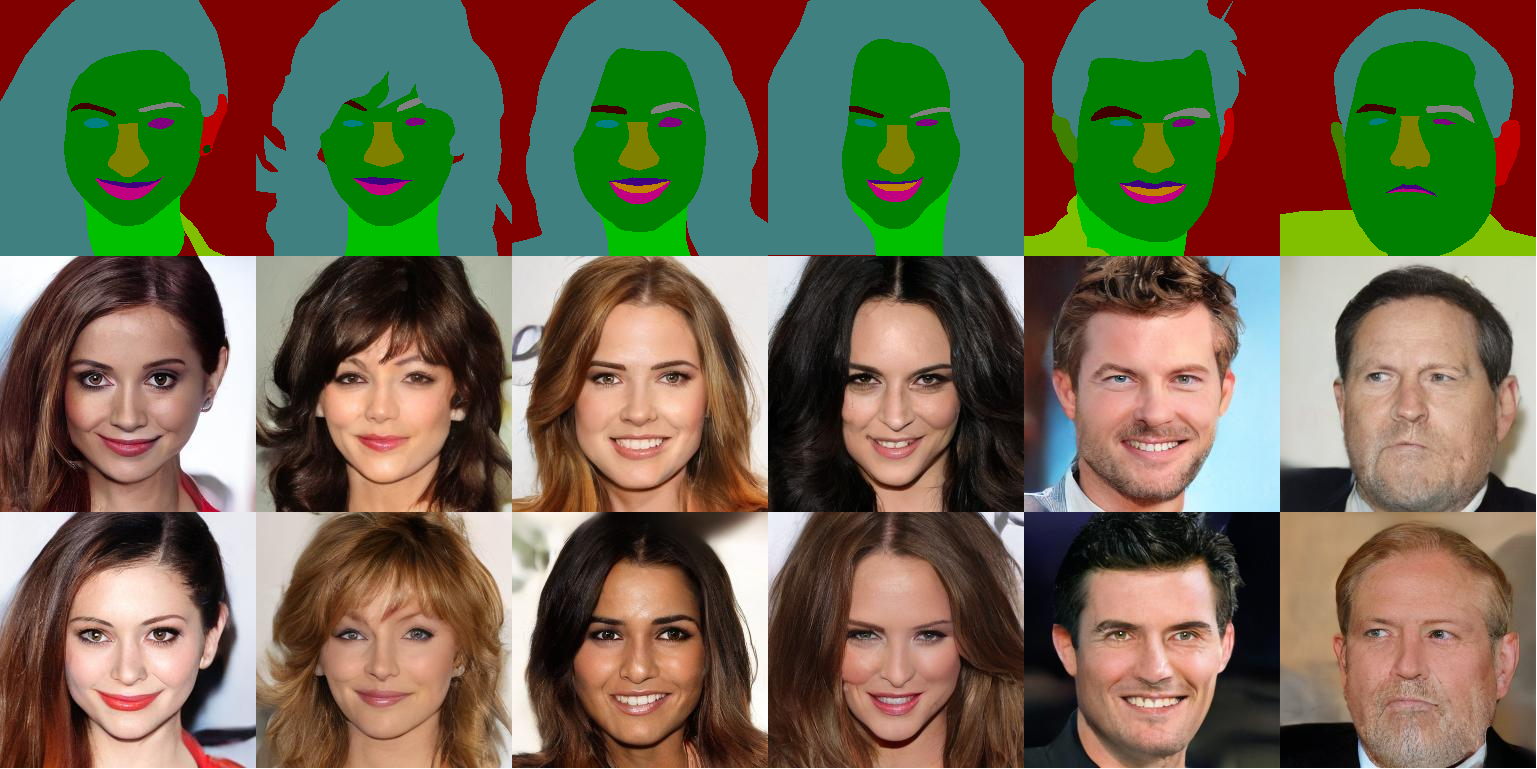}
    \caption{Results generated by changing seed to show the diversity of SISMA on CelebA.}
    \label{fig:lpips_celeba}
\end{figure}

\subsection{Qualitative Results}
In Fig. \ref{fig:celeba}, a comparison between samples generated using SDM and sample generated using \methodname{} is presented. Our model generates images with a very high quality but much more efficiently, while also preserving the semantic shape. 

Finally, as illustrated in Fig.~\ref{fig:lpips_celeba}, SISMA also has the ability to generate multiple, varied results from the same semantic mask. This exemplifies our model capability to enhance diversity and generalization which are crucial aspects for generative tasks like SIS.

\section{Conclusions}

In this paper, we proposed \methodname{}, a novel SIS architecture for human faces synthesis based on Mamba. More specifically, we substitute the standard diffusion model backbone with a Mamba-based one that is able to generate samples much more efficiently since it can be conditioned using a semantic mask without any custom normalization layer or attention layer. Experiments on several datasets and an extensive comparison with the state-of-the-art methods proved that our model is able to generate high-quality, competitive results even at a reduced computational cost.

In the future, we would like to explore the possibility of conditioning \methodname{} using reference style images in order to both control the shape and the style of the generated samples. Also, text could be added as an additional condition to our model allowing much more controllable generation.

\section{Acknowledgements}
This work was funded by ``Partenariato FAIR (Future Artificial Intelligence Research) - PE00000013, CUP J33C22002830006" funded by the European Union - NextGenerationEU through the italian MUR within NRRP, project DL-MIG.

%
%
\bibliographystyle{splncs04}
\bibliography{main}
\end{document}